# TreeMatch: A Fully Unsupervised WSD System Using Dependency Knowledge on a Specific Domain

Andrew Tran    Chris Bowes    David Brown    Ping Chen
University of Houston-Downtown

Max Choly    Wei Ding
University of Massachusetts-Boston

## Abstract

Word sense disambiguation (WSD) is one of the main challenges in Computational Linguistics. TreeMatch is a WSD system originally developed using data from SemEval 2007 Task 7 (Coarse-grained English All-words Task) that has been adapted for use in SemEval 2010 Task 17 (All-words Word Sense Disambiguation on a Specific Domain). The system is based on a fully unsupervised method using dependency knowledge drawn from a domain specific knowledge base that was built for this task. When evaluated on the task, the system precision performs above the Most Frequent Selection baseline.

## 1 Introduction

There are many words within natural languages that can have multiple meanings or senses depending on its usage. These words are called homographs. Word sense disambiguation is the process of determining which sense of a homograph is correct in a given context. Most WSD systems use supervised methods to identify senses and tend to achieve the best results. However, supervised systems rely on manually annotated training corpora. Availability of manually tagged corpora is limited and generating these corpora is costly and time consuming. With our TreeMatch system, we use a fully unsupervised domain-independent method that only requires a dictionary (WordNet, Fallbaum, 1998.) and unannotated text as input (Chen et.al, 2009).

WSD systems trained on general corpora tend to perform worse when disambiguating words from a document on a specific domain. The SemEval 2010 WSD-domain task addresses this issue by testing participant systems on documents from the environment domain. The environment domain specific corpus for this task was built from documents contributed by the European Centre for Nature Conservation (ECNC) and the World Wildlife Fund (WWF). We adapted our existing TreeMatch system from running on a general context knowledge base to one targeted at the environment domain.

This paper is organized as follows. Section 2 will detail the construction of the knowledge base. In section 3 the WSD algorithm will be explained. The construction procedure and WSD algorithm described in these two sections are similar to the procedure presented in our NAACL 2009 paper (Chen et.al, 2009). In Section 4 we present our experiments and results, and Section 5 discusses related work on WSD. Section 6 finishes the paper with conclusions.

## 2 Context Knowledge Acquisition and Representation

Figure 1 shows an overview of our context knowledge acquisition process. The collected knowledge is saved in a local knowledge base. Here are some details about each step.

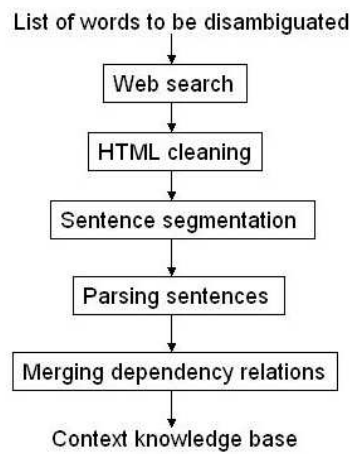

Figure 1: Context Knowledge Acquisition and Representation Process

### 2.1 Corpus Building Through Web Search

The goal of this step is to collect as many valid sample sentences as possible that contain instances of the target word. Preferably these instances are also diverse enough to contain all the different glosses of a word.

The World Wide Web is a boundless source of textual information that can be utilized for corpus building. This huge dynamic text collection represents a wide cross section of writing backgrounds that may not be represented in other corpora and may be able to better represent common human knowledge.

However, because the content on the internet is not necessarily checked for grammatical or factual accuracy, concerns may arise about the use of a corpus built from it. The quality of context knowledge will be affected by sentences of poor linguistic and poor word usage but from our experience these kind of errors are negligible when weighted against the staggering volume of valid content also retrieved.

To start the acquisition process, words that are candidates for disambiguation are compiled and saved in a text file as seeds for search queries. Each single word is submitted to a Web search engine as a query. Several search engines provide API's for research communities to automatically retrieve large number of Web pages. In our experiments we used MSN Bing! API (Bing!, 2010) to retrieve up to 1,000 Web pages and PDF documents for each to-be-disambiguated word. Collected Web pages are cleaned first, e.g., control characters and HTML tags are removed. Then sentences are segmented simply based on punctuation (e.g., ?, !, .). PDF files undergo a similar cleaning process, except that they are converted from PDF to HMTL beforehand. Sentences that contain the instances of a specific word are extracted and saved into a local repository.

### 2.2 Parsing

After the sentences have been cleaned and segmented they are sent to the dependency parser Minipar (Lin, 1998). After parsing, sentences are converted to parsing trees and saved into files. The files contain the weights of all connections between all words existing within the knowledge base. Parsing tends to take the most time in the entire WSD process. Depending on the initial size of the corpus, parsing can take weeks. The long parsing time can be attributed to Minipar's execution through system calls and also to the lack of multithreading used. However, we only need to parse the corpus once to construct the knowledge base. Any further parsing is only done on the input sentences from the words to-be-disambiguated, and the glosses of those words.

### 2.3 Merging dependency relations

After parsing, dependency relations from different sentences are merged and saved in a context knowledge base. The merging process is straightforward. A dependency relation includes one head word/node and one dependent word/node. Nodes from different dependency relations are merged into one as long as they represent the same word. An example is shown in Figure 2, which merges the following two sentences:

"Computer programmers write software."
"Many companies hire computer programmers."

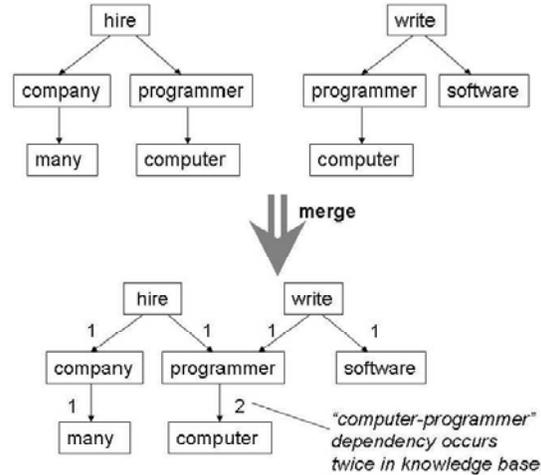

Figure 2: Merging two parsing trees. The number beside each edge is the number of occurrences of this dependency relation existing in the context knowledge base.

In a dependency relation "word1 -> word2", word1 is the head word, and word2 is the dependent word. After merging dependency relations, we will obtain a weighted directed graph with a word as a node, a dependency relation as an edge, and the number of occurrences of dependency relation as weight of an edge. This weight indicates the strength of semantic relevancy of head word and dependent word. This graph will be used in the following WSD process as our context knowledge base. As a fully automatic knowledge acquisition process, it is inevitable to include erroneous dependency relations in the knowledge base. However, since in a large text collection valid dependency relations tend to repeat far more times than invalid ones, these erroneous edges only have minimal impact on the disambiguation quality as shown in our evaluation results.

## 3 WSD Algorithm

Our WSD approach is based on the following insight:

*If a word is semantically coherent with its context, then at least one sense of this word is semantically coherent with its context.*

Assuming that the documents given are semantically coherent, if we replace a targeted to-be-disambiguated word with its glosses one by one, eventually one of the glosses will have semantic coherence within the context of its sentence. From that idea we can show the overview of our WSD procedure in Figure 3. For a given to-be-disambiguated word, its glosses from WordNet are parsed one by one along with the original sentence of the target word. The semantic coherency between the parse tree of each individual gloss and the parse tree of the original sentence are compared one by one to determine which sense is the most relevant.

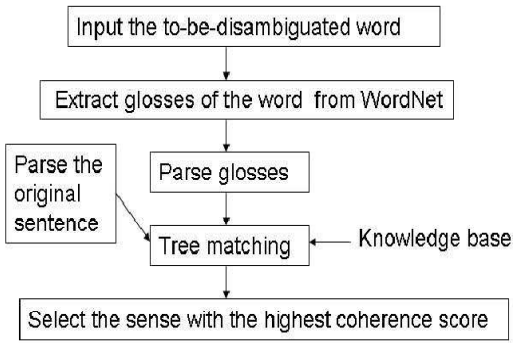

Figure 3: WSD Procedure

To measure the semantic coherence we use the following hypotheses (assume $word_1$ is the to-be-disambiguated word):

- IF in a sentence $word_1$ is dependent on $word_2$, and we denote the gloss of the correct sense of $word_1$ as $g_{1i}$, then $g_{1i}$ contains the most semantically coherent words that are dependent on $word_2$;
- In a sentence if a set of words $DEP_1$ are dependent on $word_1$, and we denote the gloss of the correct sense of $word_1$ as $g_{1i}$, then $g_{1i}$ contains the most semantically coherent words that $DEP_1$ are dependent on.

**Input: Glosses from WordNet;**
***S*: the to-be-disambiguated sentence;**
***G*: the knowledge base generated in Section 2;**
1. Input a sentence $S$, $W = \{w|\ w$'s part of speech is noun, verb, adjective, or adverb, $w \in S\}$;
2. Parse $S$ with a dependency parser, generate parsing tree $T_S$;
3. For each $w \in W$ {
4.     Input all $w$'s glosses from WordNet;
5.     For each gloss $w_i$ {
6.         Parse $w_i$, get a parsing tree $T_{wi}$;
7.         $score_d$ = TreeMatching($T_S$, $T_{wi}$);
                $Score_n$ = NodeMatching($T_S$, $T_{wi}$);
    }
8.     If the highest $score_d$ and $Score_n$ indicate the sense, choose this sense;
9.     Otherwise, choose the first sense.
10. }

**TreeMatching($T_S$, $T_{wi}$)**
11. For each node $n_{Si} \in T_S$ {
12.     Assign weight $w_{Si} = \frac{1}{lsi}$, $l_{Si}$ is the length between $n_{Si}$ and $w_i$ in $T_S$;
13. }
14. For each node $n_{wi} \in T_{wi}$ {
15.     Load its dependent words $D_{wi}$ from $G$;
16.     Assign weight $w_{wi} = \frac{1}{lwi}$, $l_{wi}$ is the level number of $n_{wi}$ in $T_{wi}$;
17.     For each $n_{Sj}$ {
18.         If $n_{Sj} \in D_{wi}$
19.         calculate connection strength $s_{ji}$ between $n_{Sj}$ and $n_{wi}$;
20.         score = score + $w_{Si} \times w_{wi} \times s_{ji}$;
21.     }
}
22. Return score;

**NodeMatching ($T_S$, $T_{wi}$)**
23. For each node $n_{Si} \in T_S$ {
24. Assign weight $w_{wi} = \frac{1}{lwi}$, $l_{wi}$ is the level number of $n_{wi}$ in $T_{wi}$;
25. For each $n_{Sj}$ {
28.     If $n_{Si} == w_{wi}$
29.     score = score + $w_{Si} \times w_{wi}$
    }
}

Figure 4: WSD Algorithm

These hypotheses are used for the functions in Figure 4. The TreeMatching function uses what we call dependency matching to ascertain the correct sense of the to-be-disambiguated word. NodeMatching function is an extension from Lesk algorithm (Lesk, 1986).

| System | Precision | Recall |
|---|---|---|
| MFS | 0.505 | 0.505 |
| TreeMatch-1 | 0.506 | 0.493 |
| TreeMatch-2 | 0.504 | 0.491 |
| TreeMatch-3 | 0.492 | 0.479 |
| Random | 0.23 | 0.23 |

Table 1: Fine-Grained SemEval 2010 Task 17 Disambiguation Scores

## 4 Experiment

The WSD-domain task for SemEval 2010 focused on the environment domain. To prepare for the tests, we constructed a new domain specific knowledge base.

Since we knew the task's domain specific corpus would be derived from ECNC and WWF materials, we produced our query list from the same source. A web crawl starting from both the ECNC and WWF main web pages was performed that retrieved 772 PDF documents. Any words that were in the PDFs and also had more than one gloss in WordNet were retained for Bing! search queries to start the acquisition process as described in section 2. 10779 unique words were obtained in this manner.

Using the 10779 unique words for search queries, the web page and PDF retrieval step took 35 days, collecting over 3 TB of raw html and PDF files, and the cleaning and sentence extraction step took 2 days, reducing it down to 3 GB of relevant sentences, while running on 5 machines. Parsing took 26 days and merging took 6 days on 9 machines. From the parse trees we obtained 2202295 total nodes with an average of 87 connections and 13 dependents per node.

Each machine was a 2.66 GHz dual core PC with 2 GB of memory with a total of 10 machines used throughout the process.

There were 3 test documents provided by the task organizers with about 6000 total words and 1398 to-be-disambiguated words. Disambiguation of the target words took 1.5 hours for each complete run. Each run used the same WSD procedure with different parameters.

The overall disambiguation results are shown in Table 1. The precision of our best submission edged out the Most Frequent Selection (MFS) baseline by .001 and is ahead of the Random selection baseline by .276.

The recall of our submissions is lower than the precision because of our reliance on Minipar for the part of speech and lemma information of the target words. Sometimes Minipar would give an incorrect lemma which at times cannot be found in WordNet and thus our system would not attempt to disambiguate the words. Previous tasks provided the lemma and part of speech for target words so we were able to bypass that step.

## 5 Related work

Generally WSD techniques can be divided into four categories (Agirre, 2006),

- Dictionary and knowledge based methods. These methods use lexical knowledge bases (LKB) such as dictionaries and thesauri, and extract knowledge from word definitions (Lesk, 1986) and relations among words/senses. Recently, several graph-based WSD methods were proposed. In these approaches, first a graph is built with senses as nodes and relations among words/senses (e.g., synonymy, antonymy) as edges, and the relations are usually acquired from a LKB (e.g., Wordnet). Then a ranking algorithm is conducted over the graph, and senses ranked the highest are assigned to the corresponding words. Different relations and ranking algorithms were experimented with these methods, such as TexRank algorithm (Mihalcea, 2005), personalized PageRank algorithm (Agirre, 2009), a two-stage searching algorithm (Navigli, 2007), Structural Semantic Interconnections algorithm (Navigli, 2005), centrality algorithms (Sinha, 2009).
- Supervised methods. A supervised method includes a training phase and a testing phase. In the training phase, a sense-annotated training corpus is required, from which syntactic and semantic features are extracted to build a classifier using machine learning techniques, such as Support Vector Machine (Novisch, 2007). In the following testing phase, the classifier picks the best sense for a word based on its surrounding words (Mihalcea, 2002). Currently supervised methods achieved the best disambiguation quality (about 80\% in precision and recall for coarse-grained WSD in the most recent WSD evaluation conference SemEval 2007 (Novisch, 2007). Nevertheless, since training corpora are manually annotated and expensive, supervised methods are often brittle due to data scarcity, and it is impractical to manually annotate huge number of words existing in a natural language.
- Semi-supervised methods. To overcome the knowledge acquisition bottleneck suffered in supervised methods, semi-supervised methods make use of a small annotated corpus as seed data in a bootstrapping process (Hearst, 1991) (Yarowsky, 1995). A word-aligned bilingual corpus can also serve as seed data (Zhong, 2009).
- Unsupervised methods. These methods acquire knowledge from unannotated raw text, and induce senses using similarity measures (Lin, 1997). Unsupervised methods overcome the problem of knowledge acquisition bottleneck, but none of existing methods can outperform the most frequent sense baseline, which makes them not useful at all in practice. The best unsupervised systems only achieved about 70% in precision and 50% in recall in the SemEval 2007 (Navigli, 2007). One recent study utilized automatically acquired dependency knowledge and achieved 73% in precision and recall (Chen, 2009), which are still below the most-frequent-sense baseline (78.89% in precision and recall in the SemEval 2007 Task 07).

Additionally there exist some "meta-disambiguation" methods that ensemble multiple disambiguation algorithms following the ideas of bagging or boosting in supervised learning (Brody, 2006).

## 6 Conclusion

This paper has described a WSD system which has been adapted for use in a specific domain for SemEval 2010 Task 17: All-Words Word Sense Disambiguation on a Specific Domain. Our system has shown that domain adaptation can be handled by unsupervised systems without the brittleness of supervised methods by utilizing readily available unannotated text from internet sources and still achieve viable results.

**Acknowledgments**

This work is partially funded by National Science Foundation grants CNS 0851984 and DHS #2009-ST-061-C10001.